%% file: main.tex
\documentclass[letterpaper]{article}

\usepackage{iccc}
\usepackage{times}
\usepackage{helvet}
\usepackage{courier}

\usepackage[square,sort,comma,numbers]{natbib}
\usepackage[utf8]{inputenc} % allow utf-8 input
\usepackage[T1]{fontenc}    % use 8-bit T1 fonts
\usepackage{CJKutf8}        % allow (some) Japanese fonts
\usepackage{url}            % simple URL typesetting
\usepackage{breakurl}
\usepackage[breaklinks]{hyperref}  % breaks url in reference.
\usepackage{hyperref}       % hyperlinks
\usepackage{booktabs}       % professional-quality tables
\usepackage{amsfonts}       % blackboard math symbols
\usepackage{nicefrac}       % compact symbols for 1/2, etc.
\usepackage{microtype}      % microtypography
\usepackage[super]{nth}
\usepackage{algorithm}
\usepackage{algorithmic}
\usepackage{wrapfig}
\usepackage{enumitem}
\usepackage{subfigure}
\usepackage{amssymb,amsmath}
\usepackage{mathtools}
\usepackage{hyperref}       % hyperlinks
\usepackage{todonotes}      % Adding visible TODO notes. See http://tug.ctan.org/macros/latex/contrib/todonotes/todonotes.pdf for docs.
\usepackage{multirow}       % For multirow cells in tables.
\usepackage{makecell}       % Line breaks inside cell in tables.

\hypersetup{
    colorlinks=true,
    linkcolor=blue,
    filecolor=magenta,      
    urlcolor=cyan,
}

\newcommand{\kk}{\textit{KaoKore }}  % The name of KaoKore. The space after is intentional.
\newcommand{\kkns}{\textit{KaoKore}}  % The name of KaoKore (no space). 
\newcommand{\fec}{\textit{Collection of Facial Expressions }}  % The space after is intentional.
\newcommand{\fecns}{\textit{Collection of Facial Expressions}}  % (no space). 
\newcommand{\ltp}{\textit{learning to paint}}
\newcommand{\ist}{\textit{intrinsic style transfer}}
\newcommand{\Ist}{\textit{Intrinsic style transfer}}
\newcommand{\Ltp}{\textit{Learning to paint}}

\title{\kkns: A Pre-modern Japanese Art Facial Expression Dataset}

\author{%
  Yingtao Tian \\
  Google Brain \\
  Tokyo, Japan
  \And
  Chikahiko Suzuki \\
  ROIS-DS Center for Open Data\\ in the Humanities\\
  NII
  \And
  Tarin Clanuwat \\
  ROIS-DS Center for Open Data\\ in the Humanities\\
  NII
  \AND
  Mikel Bober-Irizar \\
  University of Cambridge \\
  \And
  Alex Lamb \\
  MILA \\ Universit\'e de Montr\'eal \\ 
  \And
  Asanobu Kitamoto \\
  ROIS-DS Center for Open Data\\ in the Humanities\\
  NII
}

\begin{document} 
\begin{CJK}{UTF8}{min}

\maketitle

% To place the float figure* on the first page,  use the trick from http://www.tug.org/TUGboat/tb35-3/tb111beet-banner.pdf

\input{subtex/fig_emaki_example}

\begin{abstract}
From classifying handwritten digits to generating strings of text, the datasets which have received long-time focus from the machine learning community vary greatly in their subject matter.  
This has motivated a renewed interest in building datasets which are socially and culturally relevant, so that algorithmic research may have a more direct and immediate impact on society.  
One such area is in history and the humanities, where better and relevant machine learning models can accelerate research across various fields. 
To this end, newly released benchmarks ~\cite{clanuwat2018deep} and models~\cite{clanuwat2019kuronet} have been proposed for transcribing historical Japanese cursive writing, yet for the field as a whole using machine learning for historical Japanese artworks still remains largely uncharted. 
To bridge this gap, in this work we propose a new dataset \kk
% commented out for double-blind review version. For arXiv version it should be uncommeted.
\footnote{\url{https://github.com/rois-codh/kaokore}} 
which consists of faces extracted from pre-modern Japanese artwork.
We demonstrate its value as both a dataset for image classification as well as a creative and artistic dataset, which we explore using generative models. 
\end{abstract}

\section{Introduction}

In pre-modern Japan, one well-established narrative form consisted of stories displayed entirely as one long continuous painting, usually in the form of a picture scroll (絵巻物, Emakimono) or a picture book (絵本, Ehon), accompanying cursive writing of the story itself.
These stories include diverse arrays of characters (see Figure~\ref{fig:emaki_example}), and thus provide valuable materials for the study of Japanese art history.  

\enlargethispage{-18.0\baselineskip} % Part of the trick to put full width figure at the bottom of page.

In art history research, comparative style study, based on the visual comparison of characteristics in artworks, is a typical approach to answering research questions about works, such as the identification of creators, the period of production, and the skill of the painter. Among many attributes that appear in artworks, facial expressions offer especially rich information not only about the content but also about how the artworks were created. To accelerate comparative studies, \fecns~\cite{chikahiko18collection} has been created as a collection of cropped faces from multiple artworks with basic metadata annotated manually by art history experts. 
It was made possible leveraging recent technological developments, such as mass digitization of historical Japanese artworks and image sharing using IIIF~\cite{kitamono18iiif}, implemented using JSON-LD, a JSON serialization pattern for Resource Description Framework), to ensure interoperability with Linked Data.

\input{subtex/fig_fce_example}

Inspired by the recent success of developing new benchmarks \cite{clanuwat2018deep} as well as new models \cite{clanuwat2019kuronet} in the field of Japanese cursive writing,
we believe that \fec provides a largely unexplored opportunity to bridge research of historical Japanese artwork with machine learning research.
it, however, is not designed for machine learning in terms of data format (JSON-LD), image size (of different and irregular sizes) and attribute values (of both related and unrelated sets). As a result, it presents an obstacle for easy adaption of machine learning techniques.

\input{subtex/fig_dataset_example}

To bridge this gap, in this work we propose a novel dataset, \kkns, derived from the \fecns,
Our contributions can be summarized as follows:
\begin{itemize}
    \item We process the \fec to create the new \kk dataset of face images from Japanese artworks along with multiple labels for each face in a more simple, regular and easy-to-use format.  
    \item We provide standard data loaders for both PyTorch and TensorFlow as well as official train/dev/test splits which make this dataset easy to use across frameworks and compare results with.
    \item We demonstrate the dataset's utility for image classification by introducing a number of new baselines. 
    \item We study how different types of generative models can be applied to this dataset and support different types of creative exploration.
\end{itemize}

We hope that this dataset  
will help to strengthen the link between the machine learning community and research in Japanese art history.

\section{\kk Dataset}

We begin with describing \fec with is the foundation on which we build our work, the \kk dataset. \fec results from an effort by the ROIS-DS Center for Open Data in the Humanities (CODH) that has been publicly available since 2018. 
\fec provides a dataset of cropped face images extracted from Japanese artwork\footnote{Publicly available from National Institute of Japanese Literature, Kyoto University Rare Materials Digital Archive and Keio University Media Center.} from the Late Muromachi Period (16th century) to the Early Edo Period (17th century) to facilitate research into art history, especially the study of artistic style. It also provides corresponding metadata annotated by researchers with domain expertise. An example of our cropping process is shown on the left panel of Figure~\ref{fig:fce_example}.

\enlargethispage{-15.0\baselineskip} % Part of the trick to put full width figure at the bottom of page.

\input{subtex/tab_labels}

The \fec is built upon the International Image Interoperability Framework
(IIIF) and IIIF Curation Platform~\cite{kitamono18iiif}, 
which is a system designed to ease the burden of common tasks in humanities research such as viewing, searching, and annotating documents, by representing documents in a structured and machine-readable data format. 
\fec serves as the basis of several research projects in art history. One example, shown on the right panel of Figure~\ref{fig:fce_example}, is the comparison of similar-looking faces from different artworks. This type of art history research provides insights into the history of trending art styles and is useful for determining authorship of works. 

We derive \kk dataset (see Figure~\ref{fig:dataset_example} for exemplary images) from \fec in a form that will be easily recognizable to the machine learning community, with the hope of facilitating dialogue and collaboration with humanities researchers.
Concretely, we process the images and labels into industry-standard formats such that the resulting dataset is easy to use with off-the-shelf machine learning models and tools. 
Since the cropped faces from the \fec can have different sizes and aspect ratios, we pre-process them to ensure that all cropped images are normalised to the same size and aspect ratio.  

The resulting \kk dataset contains $5552$ RGB image files of size $256$ x $256$.  Figure~\ref{fig:dataset_example} shows examples from the \kk dataset demonstrate its collection of various faces in diverse yet coherent artistic styles.
The format of images follows that of ImageNet~\cite{imagenet_cvpr09},
making it a potential drop-in replacement dataset for existing unsupervised  learning setups.  

To facilitate supervised learning, we provide two sets of labels for all faces: \textit{gender} and \textit{(social) status}, which belong to the frequently appearing subset of all expert-annotated labels from the \fecns.
Table~\ref{tab:labels} shows the classes and labels available in the \kk dataset, with exemplary images belonging to each label.
Note that  we setup tasks on the labels that appear most frequently. For example for class \textit{(social) status} we choose \textit{noble}, \textit{warrior}, \textit{incarnation} and \textit{commoner} which each has at least 600 images, while discarding rare ones like \textit{priesthood} and \textit{animal} which each has merely a dozen. This is to avoid an overly unbalanced distribution over the labels.
We also give official training, validation, and test sets splits to enable straightforward model comparisons in future studies.

\section{Experiments}

We conduct two types of experiments.
First, we provide \textit{quantitative results} on the supervised machine learning tasks of gender and status classification of \kk images.
Second, we provide \textit{qualitative results} from generative models on the \kk images.

\subsection{Classification Results for \kk Dataset}

\input{subtex/tab_classification_accuracies}

\input{subtex/fig_media-stylegan}

We present classification results for the \kk dataset in Table~\ref{tab:classification_accuracies}
using several neural network architectures,
namely VGG~\cite{simonyan2014very}, AlexNet~\cite{krizhevsky2014one}, 
ResNet~\cite{he2016deep},  MobileNet-V2~\cite{sandler2018mobilenetv2}, DenseNet~\cite{huang2017densely} and Inception~\cite{szegedy2016rethinking}.
We use PyTorch’s reference implementation\cite{pytorch}
of common visual models, standard data augmentation as well as the Adam optimizer\cite{kingma2014adam}.  
We use early stopping on the validation set, and report the test set accuracy.
% Training setup could be found in Appendix\todo{fill the appendix for training details}.

\input{subtex/fig_intrinsic-style-transfer-steps}

In providing accuracies across various models we demonstrate that standard classifiers are able to achieve decent but imperfect performance on these tasks.  Additionally, we show that newer and larger architectures often achieve better performance, which suggests that even further improvement through better architectures will be possible.%  \todo{maybe also provide failure case for these models}

\enlargethispage{-20.5\baselineskip} % Part of the trick to put full width figure at the bottom of page.

\subsection{Creativity Applications}

\input{subtex/fig_intrinsic-style-transfer-finals}

As the \kk dataset is based on artwork, we also investigate its creative applications.  While our hope is that people will find novel ways of engaging with this dataset artistically, we demonstrate that reasonable results can be achieved by using today's best-performing generative models.  We note that faces in the \kk dataset contain many correlated attributes like face shape and hairstyle, giving a challenging tasks of correctly model correlations, yet are highly stylized and much simpler than realistic images of faces, leading to a more challenging support of data distribution. Since these two challenges are non-trivial problems for generative models, we hope the \kk dataset will be useful for generative modeling research.

\input{subtex/fig_intrinsic-style-transfers-many}

\input{subtex/fig_learning-to-paint-steps}

\subsubsection{Generative Adversarial Networks}

We first explore Generative Adversarial Networks (GANs)~\cite{goodfellow2014generative} which have been successfully used as generative models for synthesizing high quality images\cite{karras2017progressive,karras2019style,zhang2018self},
and has also seen creative applications such as image-to-image translation~\cite{zhu2017unpaired}, controllable Anime character generation~\cite{jin2017towards,hamada2018full,crypko}, photo-realistic face generation~\cite{lu2018attribute},  apartment~\cite{archigan}, and fashion design~\cite{chen2020tailorgan}.

Inspired by this, we leverage StyleGAN~\cite{karras2019style}, a state-of-the-art GAN model for images.
We implement and train it on our dataset and show the resulting generated images.
In Figure~\ref{fig:media-stylegan}, we show uncurated images produced by StyleGAN,
showing that the varieties in our datasets are successfully captured by the generative model.

\subsubsection{Neural Painting Models}

\enlargethispage{-21.5\baselineskip} % Part of the trick to put full width figure at the bottom of page.

\input{subtex/fig_learning-to-paint-finals}

\input{subtex/fig_learning-to-paint-many}

We find that GAN models are able to generate somewhat plausible-looking images of \kk faces (Figure~\ref{fig:media-stylegan}).  
However, the GAN objective requires that the model directly generate \textit{pixels}, where an artist would paint the image by applying strokes iteratively on a canvas.  Thus when a GAN makes mistakes, the types of errors it makes are generally quite unlike the variations that could be produced by a human painter.  To give the synthesis process a more artwork-like inductive bias, we consider
\textit{Stroke-based rendering}~\cite{hertzmann2003survey}
which produces a reference painting by sequentially drawing primitives, such as simple strokes, onto a canvas. 
Recent advances using neural networks have been proposed by integrating techniques such as differentiable image parameterizations \cite{mordvintsev2018differentiable,nakano2019neural} or reinforcement learning\cite{pmlr-v80-ganin18a,huang2019learning},
which can greatly improve the quality of generated painting sequences.  
Given that our proposed \kk dataset is based on art collections,
we are interested in applying these neural painting methods with image production mechanisms that better resemble human artists.
In particular, we explore applying \ist~\cite{nakano2019neural} and \ltp~\cite{huang2019learning} on the proposed \kk dataset.

\subsubsection{\Ist~\cite{nakano2019neural}} is a set of methods that combines differentiable approximation with non-differentiable painting primitives (e.g., strokes in our colored pencil drawing style) and an adversarially trained neural agent that learns to recreate the reference image on the canvas by producing these primitives sequentially.
It is characterized by a lack of ``style-loss'' that are often leveraged in style transfer methods that carries the reference image's style into the target one, which in turn exposes the intrinsic style derived from painting primitives mentioned above. 
In Figure~\ref{fig:intrinsic-style-transfer-steps}, we show the produced painting sequences on a few exemplary images.
It can be observed that the image has been decomposed into strokes that resemble how a human artist might create the pencil sketch, while the model has not been provided with any recording of sketching sequence.
This is further epitomized in Figure~\ref{fig:intrinsic-style-transfer-finals} and Figure~\ref{fig:intrinsic-style-transfers-many}, which show the reference images and the final canvas after applying complete painting.

\subsubsection{\Ltp~\cite{huang2019learning}} is a neural painting model which differentiates itself from others in a few aspects,
including using regions marked by quadratic B\'ezier curves as painting primitives
as well as leveraging model-based deep reinforcement learning for training.
As a result, its painting sequence is made of simple regions rather than brush-like strokes,
and the sequence is as short as possible due to the optimization goal used in reinforcement learning training. 
As shown in Figure~\ref{fig:learning-to-paint-steps}, Figure~\ref{fig:learning-to-paint-finals} and Figure~\ref{fig:learning-to-paint-many}
the method learns to assemble simple curve regions in recreating the image that emphasize the general arrangements of objects in the painting and resembles an abstract painting style. 

The two neural painting models explored can, given a single image from the \kk dataset, produce painting sequences that could be applied on a (virtual) canvas and resemble human-interpretable styles. Yet due to each method's fundamentally different mechanism, the style, while being expressive, as a result also resort to different artistic style. By simultaneously presenting artistic familiarity in the style and surprising in how to decompose a single image, 
we hope this result can provide insights into the study of art styles.

\section{Discussion and Future Work}

We hope that our proposed \kk dataset provides a foundation on which future works can be built: including humanities research of historical Japanese artworks or machine learning research with creative models, given our dataset's dual value both as a  classification dataset as well as a creative and artistic one.
In the future, we plan to increase the number of facial expression images in our dataset by building a machine learning powered human-in-the-loop annotation mechanism that allows us to scale the labeling process.
We would also like to construct new datasets in future work which help to expand machine learning research and its applications for Japanese art, beyond the face images considered in the \kk dataset.
Finally, we anticipate that further interdisciplinary collaboration between humanities research and the machine learning research community would contribute to better and more efficient cultural preservation. 

\bibliographystyle{iccc}
\bibliography{bib}

\end{CJK}
\end{document}

%% file: subtex/fig_emaki_example.tex
\begin{figure}[b!]
    \setlength{\hfuzz}{1.1\columnwidth}
    \vskip -0.5\baselineskip % could be tuned
    \begin{minipage}{\textwidth}
        \begin{center}
            \includegraphics[trim={5cm 25cm 0 35cm},clip,width=1.0\textwidth]{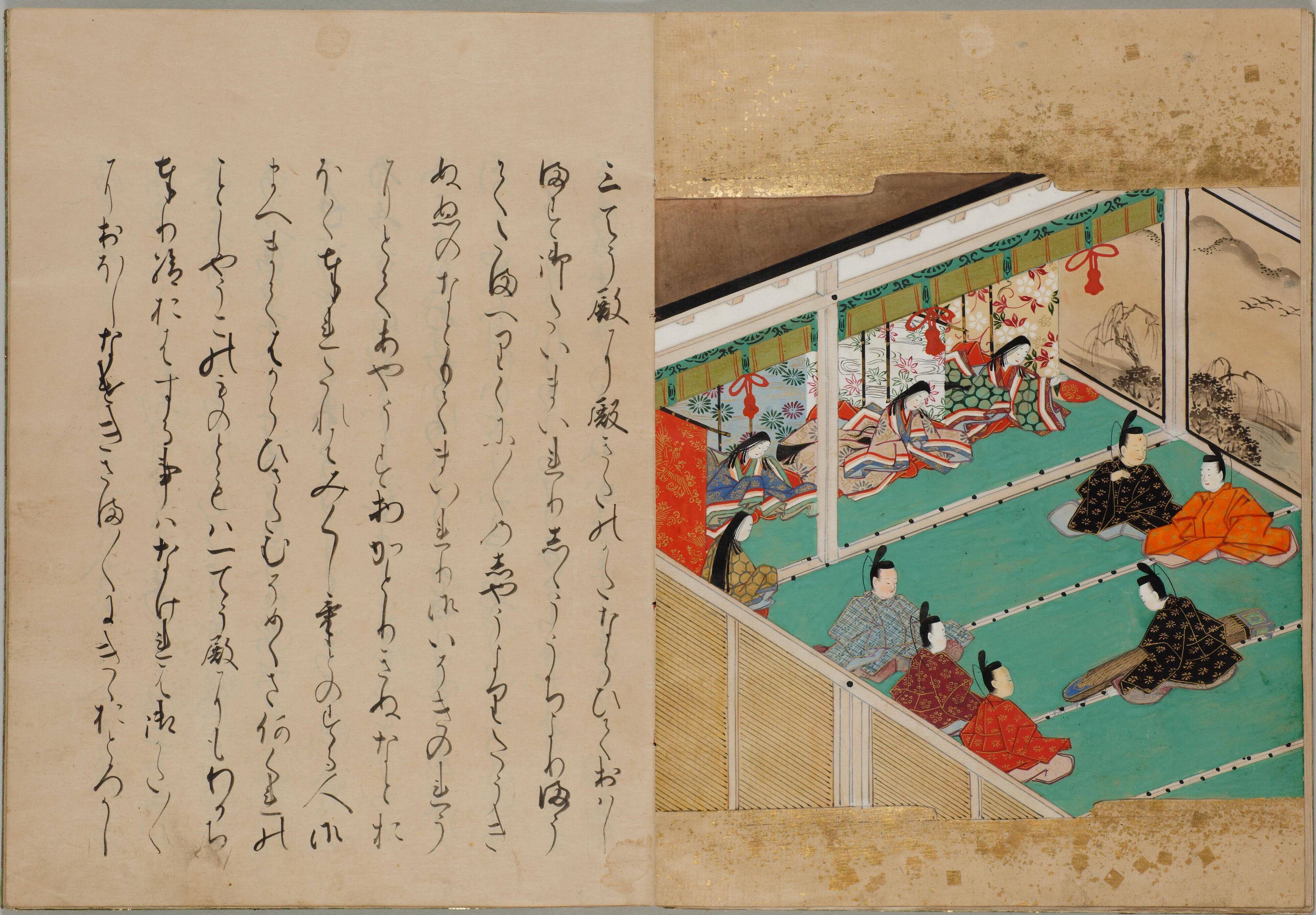}
                            % trim={<left> <lower> <right> <upper>} . 
            \vskip -0.7\baselineskip % could be tuned
            \caption{A cropped page of \textit{Tale of the Hollow Tree} (宇津保物語, Utsuho Monogatari), a late-10th century Japanese story, presented in the form of a picture book (絵本, Ehon) in the 17th century~\cite{utsuho}. 
            Picture scrolls and picture books usually have cursive texts telling the story (left) in addition to story-explaining paintings depicting many characters (right), often on the same page.
            }
            \label{fig:emaki_example}
        \end{center}
    \end{minipage}
    \vskip -0.5\baselineskip % could be tuned
\end{figure}

%% file: subtex/fig_fce_example.tex
\begin{figure*}[!t]
    \begin{center}
        \centerline{
            \includegraphics[width=0.30\textwidth]{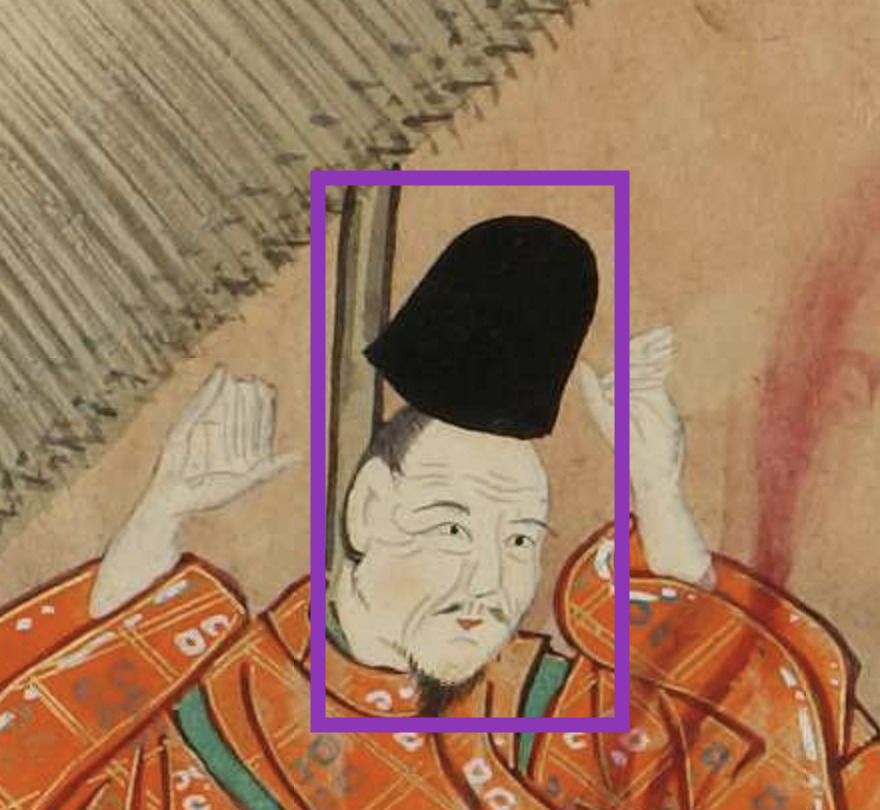}
            \hspace{5mm}
            \includegraphics[trim={0 0.5cm 0 0},clip,width=0.65\textwidth]{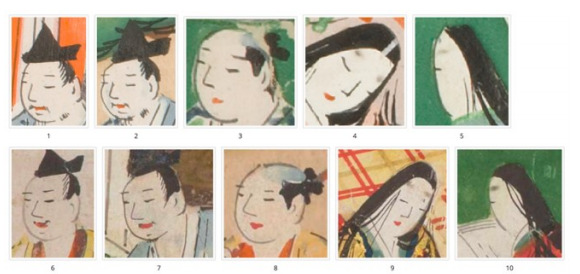}
        }
        \vskip -0.7\baselineskip % tune it later
        \caption{On the left is a face in the \fec showing one crop with annotated metadata ``Gender: Male (性別：男)'' and ``Status: Nobleman (身分：貴族)''. 
        On the right is an example of style comparison using the \fec, displaying similar faces from two different works~\cite{chikahiko18collection}.}
        \label{fig:fce_example}
    \end{center}
    \vskip -0.7\baselineskip % could be tuned
\end{figure*}

%% file: subtex/fig_dataset_example.tex
\begin{figure}[!b]
    \setlength{\hfuzz}{1.1\columnwidth}
    \begin{minipage}{\textwidth}
        \begin{center}
            \includegraphics[width=1.0\textwidth]{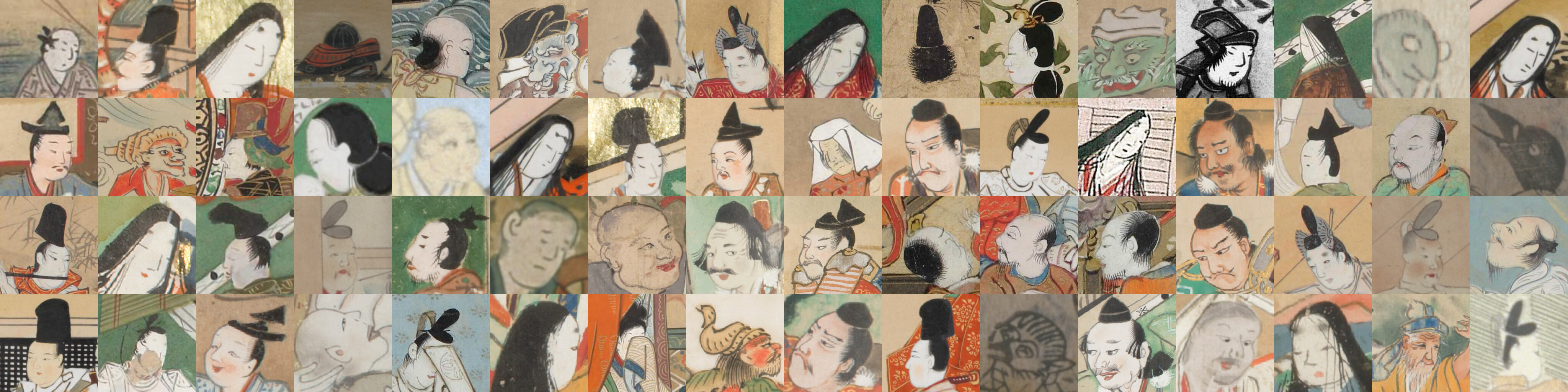}
            \vskip -0.7\baselineskip % tune it later
            \caption{Exemplary images in the \kk dataset. These examples demonstrate various faces dipicting different subjects in diverse yet coherent artistic styles.
            }
            \label{fig:dataset_example}
        \end{center}
    \end{minipage}
\end{figure}

%% file: subtex/tab_labels.tex
\begin{table*}[t]
    \begin{center}
        \begin{small}
            \begin{tabular}{ccp{3in}}
                \toprule
                \textit{Class} & \textit{Labels}   & \textit{Examples} \\        
                \midrule
                \makecell{gender\\(性別)} & 
                        \makecell{male\\(男)} & 
                        \parbox[c]{1em}{
      \includegraphics[width=3in]{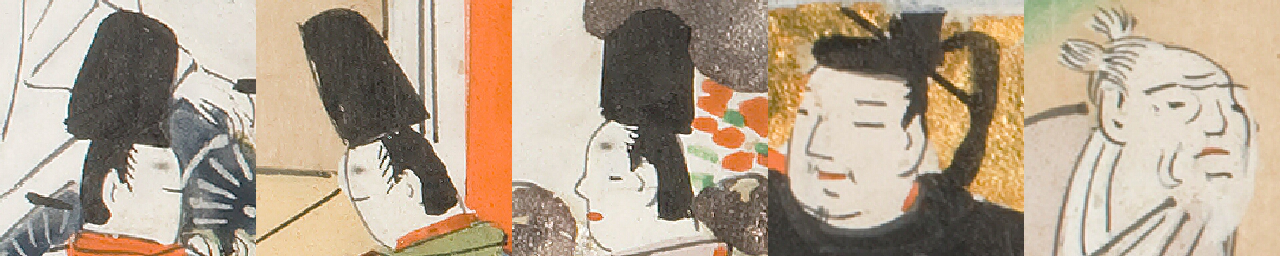}} \\
                \cmidrule{2-3}
                        & 
                        \makecell{female\\(女)} & 
                        \parbox[c]{1em}{
      \includegraphics[width=3in]{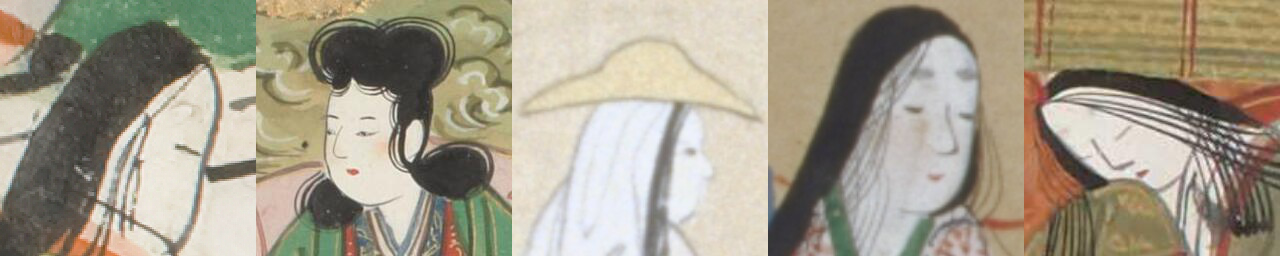}} \\
                \midrule
                \makecell{status\\(身分)} &  \makecell{noble\\(貴族)}            & \parbox[c]{1em}{
      \includegraphics[width=3in]{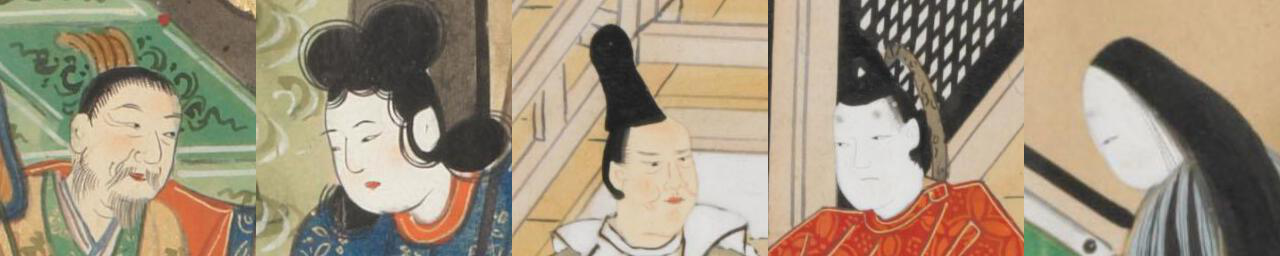}} \\
                \cmidrule{2-3}
                              &  \makecell{warrior\\(武士)}     & \parbox[c]{1em}{
      \includegraphics[width=3in]{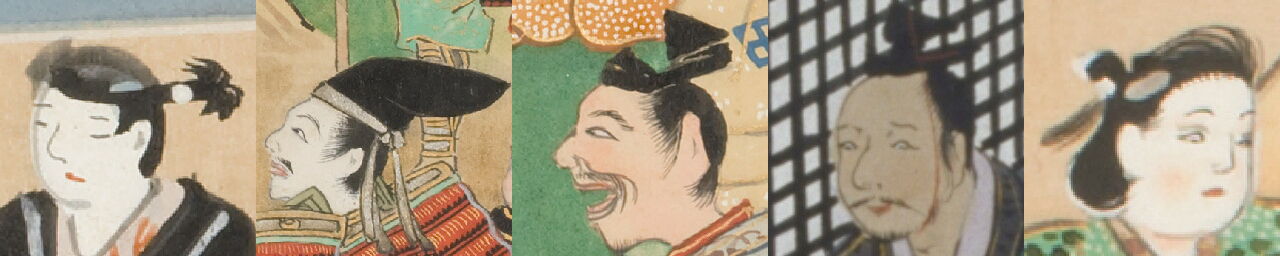}}  \\
                \cmidrule{2-3}
                              &  \makecell{incarnation\\(化身)} & \parbox[c]{1em}{
      \includegraphics[width=3in]{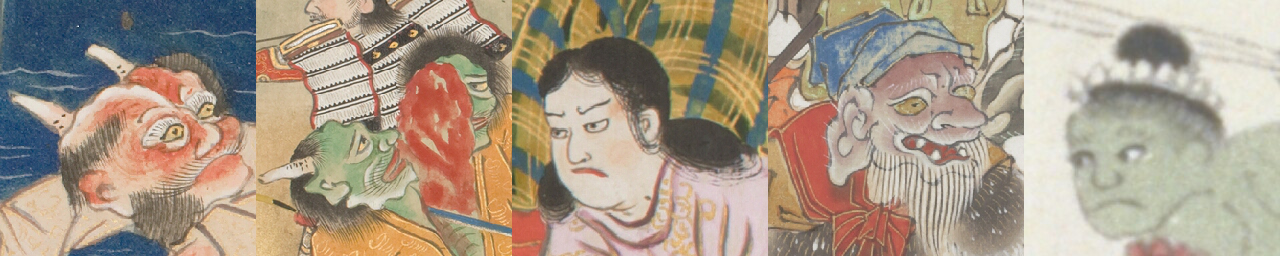}}\\
                \cmidrule{2-3}
                              &  \makecell{commoner\\(庶民)}    & \parbox[c]{1em}{
      \includegraphics[width=3in]{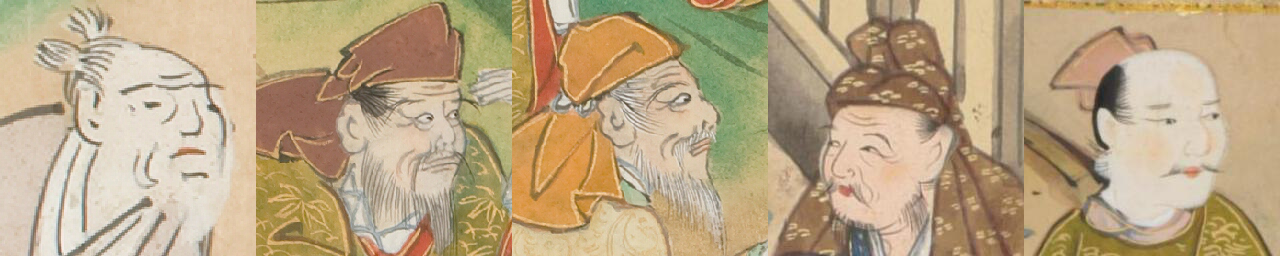}}\\
                \bottomrule
            \end{tabular}
        \end{small}
    \end{center}
    \vskip -0.5\baselineskip % tune it later
    \caption{Labels available in the dataset along with exemplary images belonging to each label.}
    \vskip -0.5\baselineskip % tune it later
    \label{tab:labels}
\end{table*}

%% file: subtex/tab_classification_accuracies.tex
\begin{table}[!h]
    \begin{center}
        \begin{small}
            \begin{tabular}{lcc}
                \toprule
                Model & classifying \textit{gender} & classifying \textit{status} \\
                \midrule
                VGG-11~\cite{simonyan2014very} & $92.03$ \% & $78.74$ \% \\
                AlexNet~\cite{krizhevsky2014one} & $91.27$ \% & $ 78.93$ \% \\
                ResNet-18~\cite{he2016deep} & $92.98$ \% & $82.16$ \% \\
                ResNet-34~\cite{he2016deep} & $93.55$ \% & $84.82$ \% \\
                MobileNet-v2~\cite{sandler2018mobilenetv2} & $95.06$ \% & $82.35$ \% \\
                DenseNet-121~\cite{huang2017densely} & $94.31$ \% & $79.70$ \% \\
                Inception-v3~\cite{szegedy2016rethinking} & $96.58$ \% & $84.25$ \% \\
                \bottomrule
            \end{tabular}
        \end{small}
    \end{center}
    \vskip -0.5\baselineskip % tune it later
    \caption{Test accuracy on classification tasks.}
    \label{tab:classification_accuracies}
\end{table}

%% file: subtex/fig_media-stylegan.tex
\begin{figure*}[!ht]
    \begin{center}
        \centerline{\includegraphics[width=1.0\textwidth]{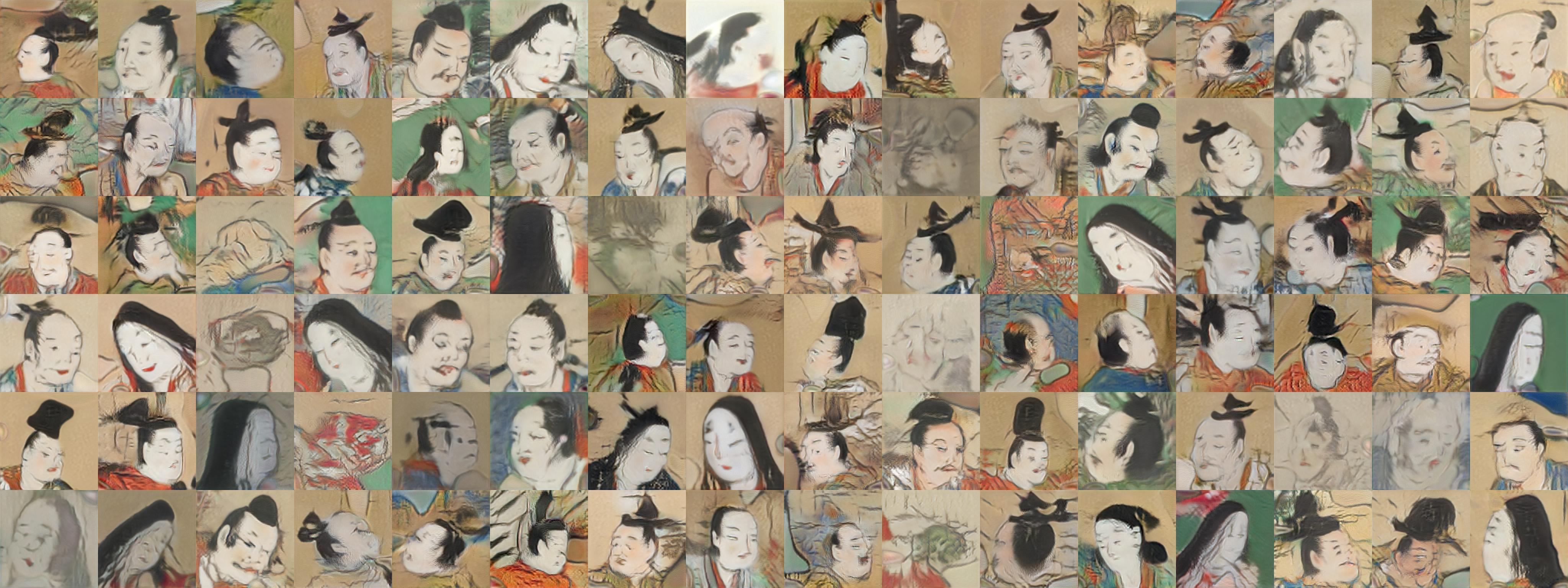}}
        \caption{Uncurated images produced by StyleGAN~\cite{karras2019style} trained on our dataset. These samples demonstrate that the varieties in our datasets is well captured.}
        \label{fig:media-stylegan}
    \end{center}
    \vskip -1.5\baselineskip % could be tuned
\end{figure*}

%% file: subtex/fig_intrinsic-style-transfer-steps.tex
\begin{figure}[b!]
    \setlength{\hfuzz}{1.1\columnwidth}
    \begin{minipage}{\textwidth}
    \begin{center}
        \hrule
        \centerline{\includegraphics[width=1.0\textwidth]{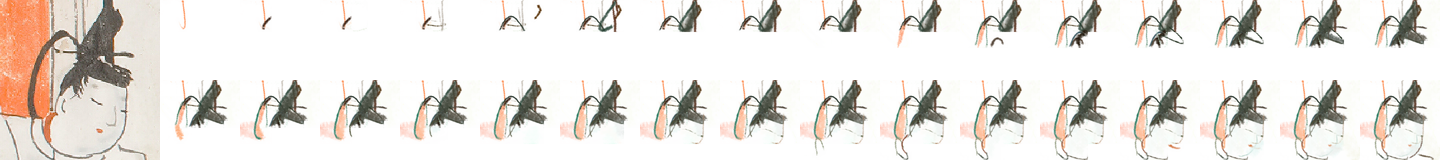}}
        \hrule
        \centerline{\includegraphics[width=1.0\textwidth]{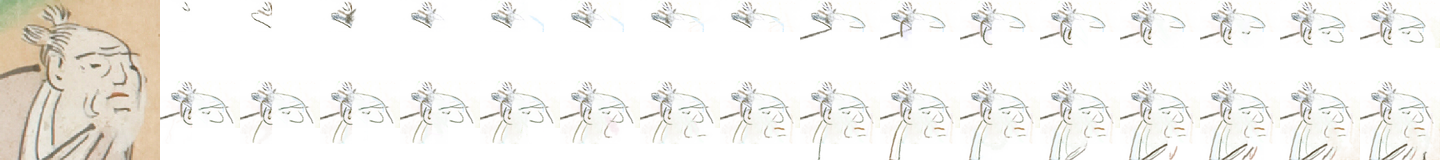}}
        \hrule
        \centerline{\includegraphics[width=1.0\textwidth]{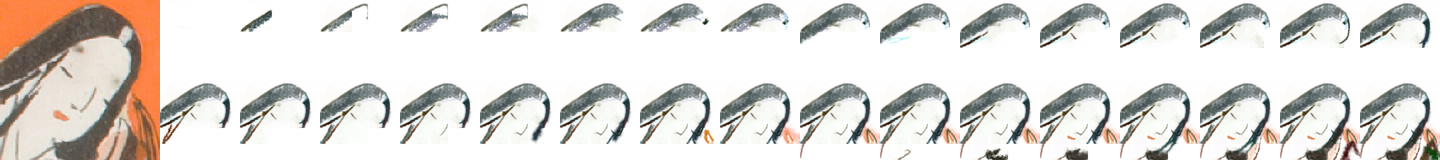}}
        \hrule
        \caption{Painting sequences generated by \ist~\cite{nakano2019neural}, a neural painting model, on a few exemplary images in the \kk dataset. On each row, the leftmost image is the reference image from \kk dataset, while the smaller images illustrate the generate painting sequence.}
        \label{fig:intrinsic-style-transfer-steps}
    \end{center}
    \end{minipage}
    \vskip -1.0\baselineskip % could be tuned
\end{figure}

%% file: subtex/fig_intrinsic-style-transfer-finals.tex
\begin{figure*}[t!]
    \begin{center}
        \centerline{\includegraphics[width=1.0\textwidth]{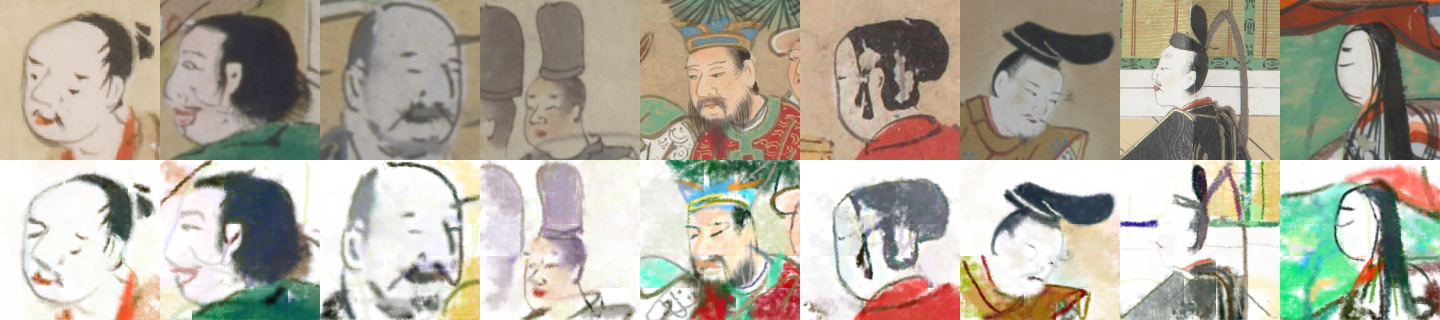}}
        \caption{The second row shows the final produced canvas from \ist~\cite{nakano2019neural} after all painting steps; trained to approximate the reference image (first row).}
        \label{fig:intrinsic-style-transfer-finals}
    \end{center}
    \vskip -2.0\baselineskip % could be tuned
\end{figure*}

%% file: subtex/fig_intrinsic-style-transfers-many.tex
\begin{figure}[h]
    \begin{center}
        \centerline{\includegraphics[width=0.8\columnwidth]{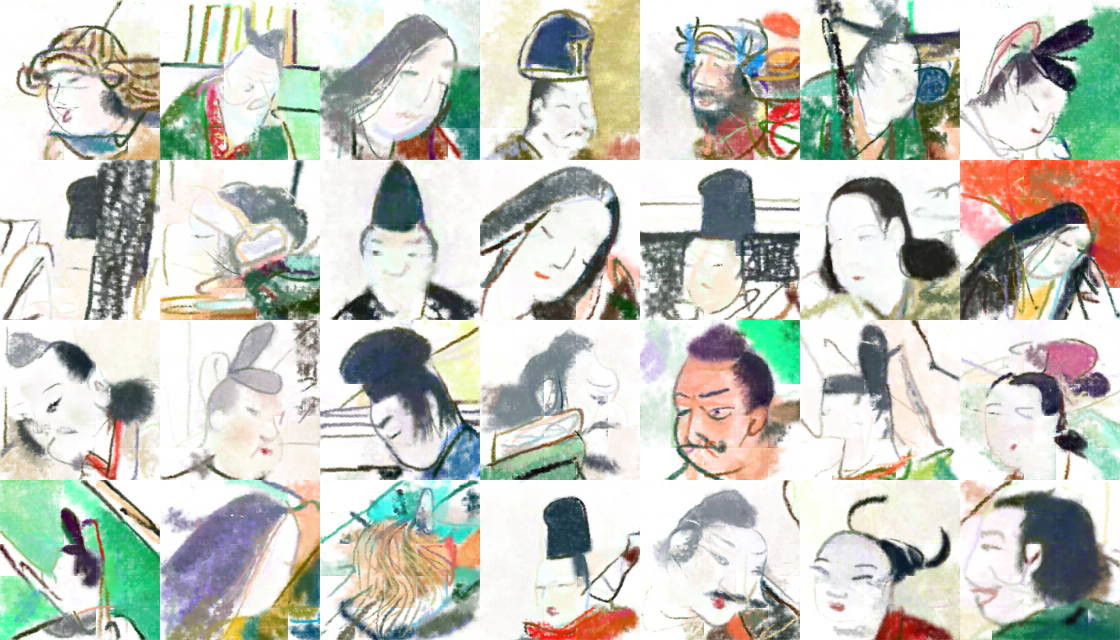}}
        \caption{\ist~\cite{nakano2019neural} produces more example of final canvas after all painting steps are generated.} 
        \label{fig:intrinsic-style-transfers-many}
    \end{center}
    \vskip -1.0\baselineskip % tune it later
\end{figure}

%% file: subtex/fig_learning-to-paint-steps.tex
\begin{figure}[b!]
    \setlength{\hfuzz}{1.1\columnwidth}
    \begin{minipage}{\textwidth}
    \begin{center}
        \hrule
        \centerline{\includegraphics[width=1.0\textwidth]{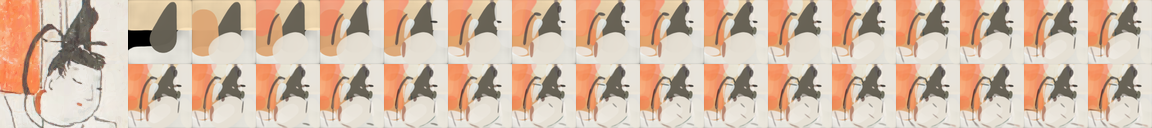}}
        \hrule
        \centerline{\includegraphics[width=1.0\textwidth]{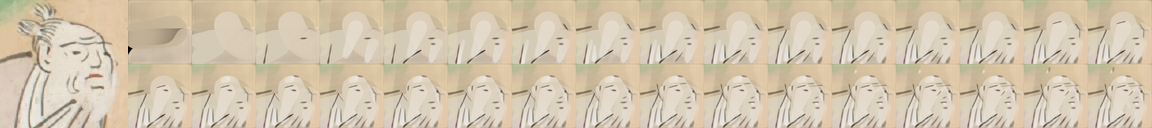}}
        \hrule
        \centerline{\includegraphics[width=1.0\textwidth]{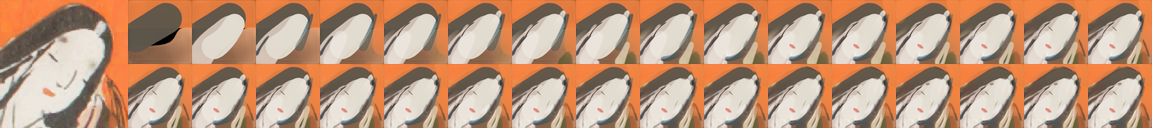}}
        \hrule
        \caption{Painting sequences produced by \ltp~\cite{huang2019learning}, a neural painting model, on a few exemplary images in the \kk dataset. On each row, the leftmost image is the reference image from the\kk dataset, while the smaller images illustrate the generated painting sequence. The reference images used are the same as in Figure~\ref{fig:intrinsic-style-transfer-steps} for easier comparison of style studies.}
        \label{fig:learning-to-paint-steps}
    \end{center}
    \end{minipage}
    \vskip -1.0\baselineskip % could be tuned
\end{figure}

%% file: subtex/fig_learning-to-paint-finals.tex
\begin{figure*}[!t]
    \begin{center}
        \centerline{\includegraphics[width=1.0\textwidth]{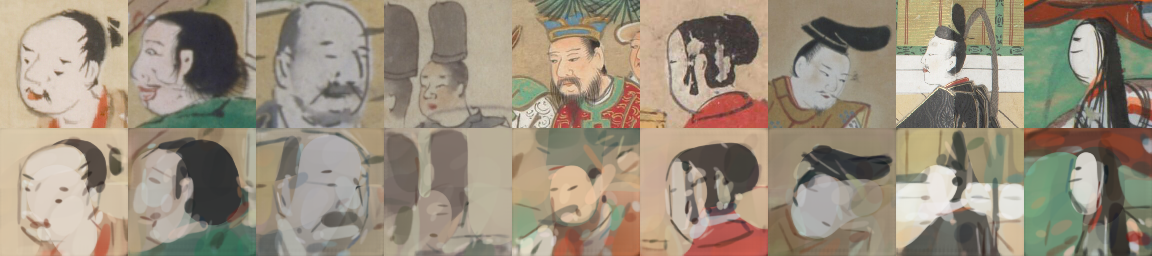}}
        \caption{Final canvases produced by \ltp~\cite{huang2019learning} (second row) after all painting steps have been completed in order to approximate the reference image (first row). Reference images match Figure~\ref{fig:intrinsic-style-transfer-finals} for easier comparison of style studies.}
        \label{fig:learning-to-paint-finals}
    \end{center}
    \vskip -1.0\baselineskip % could be tuned
\end{figure*}

%% file: subtex/fig_learning-to-paint-many.tex
\begin{figure}[h]
    \begin{center}
        \centerline{\includegraphics[width=0.8\columnwidth]{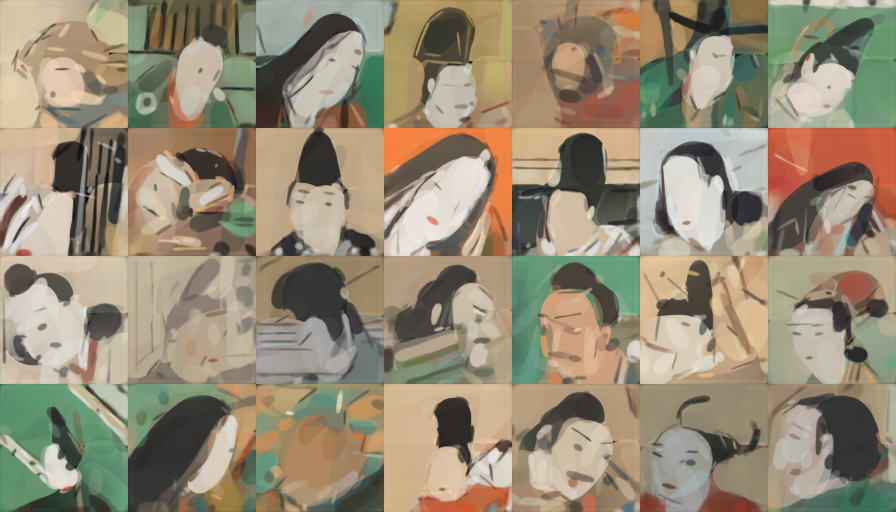}}
        \caption{More examples of \ltp~\cite{huang2019learning} on \kk images after all painting steps. Reference images match Figure~\ref{fig:intrinsic-style-transfers-many} for easier comparison of style studies.}
        \label{fig:learning-to-paint-many}
    \end{center}
    \vskip -1.0\baselineskip % could be tuned
\end{figure}